# Phonotactic Complexity across Dialects

Ryan Soh-Eun Shim*, Kalvin Chang*, David R. Mortensen
Institute for Natural Language Processing, University of Stuttgart
Language Technologies Institute, Carnegie Mellon University
soh-eun.shim@ims.uni-stuttgart.de
{kalvinc, dmortens}@cs.cmu.edu

## Abstract

Received wisdom in linguistic typology holds that if the structure of a language becomes more complex in one dimension, it will simplify in another, building on the assumption that all languages are equally complex (Joseph and Newmeyer, 2012). We study this claim on a micro-level, using a tightly-controlled sample of Dutch dialects (across 366 collection sites) and Min dialects (across 60 sites), which enables a more fair comparison across varieties. Even at the dialect level, we find empirical evidence for a tradeoff between word length and a computational measure of phonotactic complexity from a LSTM-based phone-level language model—a result previously documented only at the language level. A generalized additive model (GAM) shows that dialects with low phonotactic complexity concentrate around the capital regions, which we hypothesize to correspond to prior hypotheses that language varieties of greater or more diverse populations show reduced phonotactic complexity. We also experiment with incorporating the auxiliary task of predicting syllable constituency, but do not find an increase in the negative correlation observed.

## 1. Introduction

Phonotactics refers to a set of language- or dialect-specific constraints on what constitutes a licit or illicit sound sequence. Phonotactic complexity refers to the variety of structures allowed at different positions within a syllable or word—essentially how unpredictable a language variety's phonemes behave at different positions. If we view the possible upcoming phonemes as branches of a tree structure, where the root denotes the start of the word, the phonotactic complexity is the size of the tree. For instance, a language that requires VCV sequences to obey vowel harmony has lower phonotactic complexity than another VCV-language without this phonotactic constraint, since the rule reduces the number of possible vowels in each position (Pimentel et al., 2020).

Recent results have shown a moderate negative correlation between phonotactic complexity and average word length in a dataset of 1,016 basic concept words across 106 languages (Pimentel et al., 2020). A decrease in word length accompanies an increase in phonotactic complexity, which the compensation hypothesis (Hockett, 1958; Moran and Blasi, 2014) would argue occurs as a compensatory mechanism. However, there has been comparably little work that examines this hypothesis in the context of dialects of the same language. Such research would provide a microcosmic view of complexity where most variables (e.g., areal influences, phylogenetic biases, variation in descriptive conventions) are held relatively constant. This addresses the problem of typological imbalance in Pimentel et al. (2020)'s study, where the NorthEuraLex dataset they use (Dellert et al., 2020) favors Uralic and Indo-European languages. Examining dialects additionally holds much interest for dialectology, since geographical effects such as spatial autocorrelation can be observed, and areas of low or high linguistic complexity (in some dimension) can be explained. For example, geographic and demographic factors such as population size may (inversely) correlate with complexity (Lupyan and Dale, 2010; Dale and Lupyan, 2012). Indeed, this is what we find.[1]

We measure the phonotactic complexity of a dialect with Shannon entropy (bits per phoneme), estimated with Pimentel et al. (2020)'s LSTM-based phonotactic language model (Equation 1). While predicting the next phoneme accounts for the number of options at some position, their model only considers the order of segments in a word and fails to account for the interaction between syllable structure and phonotactic constraints (subsection 4.2). Our contributions are as follows:

1. **Phonotactic tradeoffs across dialect continua**: We corroborate the negative corre-

---

*These authors contributed equally.

[1] Recently, Shcherbakova et al. (2023) do not find such an inverse correlation in a dataset of 1,314 languages (Skirgård et al., 2023), but their study examines different dimensions of linguistic complexity.

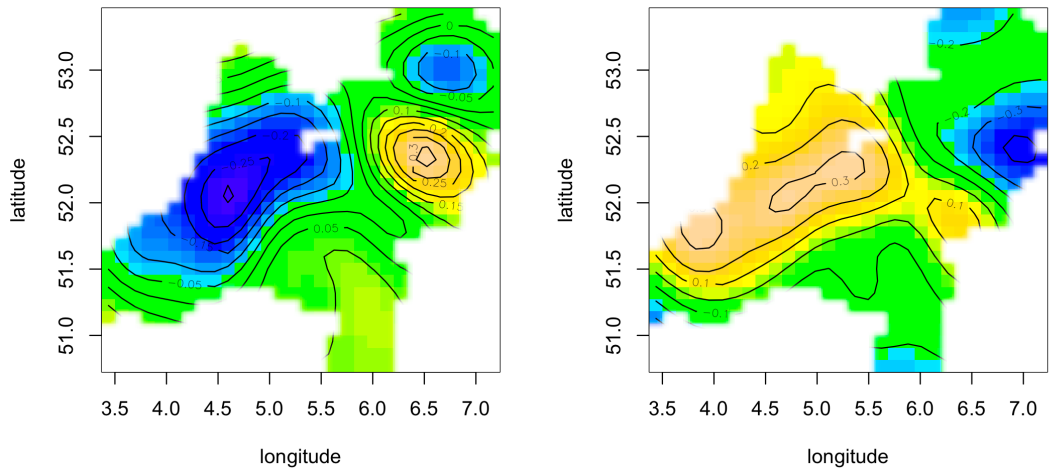

Figure 1: Contour plot of GAM results using a thin plate regression spline, where the response variable is phonotactic complexity (left) and average word length (right) of Dutch dialects, and the predictors, their latitude and longitude. Colder colors (blue) indicate low complexity, while warmer colors (yellow) indicate higher values.

lation between word length and phonotactic complexity found in Pimentel et al. (2020) even when examined in our controlled setting of a large number of very similar language varieties.

2. **Social explanation of complexity distributions**: Areas of low phonotactic complexity concentrate around the capital regions, which supports the hypothesis that varieties with more diverse populations tend to have simpler phonology and morphology, potentially due to learnability constraints (Lupyan and Dale, 2010; Dale and Lupyan, 2012; Bentz and Winter, 2014).

3. **Supervised syllable constituency prediction**: We incorporate syllabification into a phonotactic language model by virtue of multi-task learning, and observe that knowledge of syllable constituency does not increase the negative correlation in our data.

## 2. Related Work

Pimentel et al. (2020) are not the first to model phonotactics computationally. Hayes and Wilson (2008) use a maximum entropy grammar to assign scores of well-formedness to sequences of phonemes based on how many phonotactic constraints are violated. The model learns *SPE*-style (Chomsky and Halle, 1968) constraints from primary data.

Mayer and Nelson (2020) find that RNN-based phoneme-level language models can learn challenging phonotactic patterns, such as Finnish vowel harmony (a long-distance pattern). While they train on articulatory feature vectors, they do not incorporate syllable structure into their model. They hypothesize that an Elman RNN may suffice for the task of phonotactic modeling (though we use an LSTM).

Using Mayer and Nelson (2020)'s model, Kirby (2021a) finds that the phonotactic probability of a syllable is not affected by the placement of tone in the phonetic transcription. They do not train on transcriptions of words and instead train on all possible syllables for 4 languages with syllables as tone-bearing units (Mandarin, Thai, Vietnamese, Cantonese) to maximize the probability of the syllable lexicon.

Steuer et al. (2023) find that LSTM-based phoneme-level language models can capture vowel harmony, a phonotactic constraint, in different languages. They quantify the degree of vowel harmony with an entropy-based measure derived from the likelihood but revise the language model objective to only predict the next vowel. In contrast, the language model we use is not restricted to a single phonotactic constraint.

Daland et al. (2011a) use a version of Hayes and Wilson (2008)'s phonotactic model supervised with syllable structure and find that models need to incorporate both sonority and syllabification to capture the phonotactic phenomenon of sonority projection. We implicitly incorporate sonority by incorporating syllable constituency obtained via sonority-based syllabification (subsection 4.3).

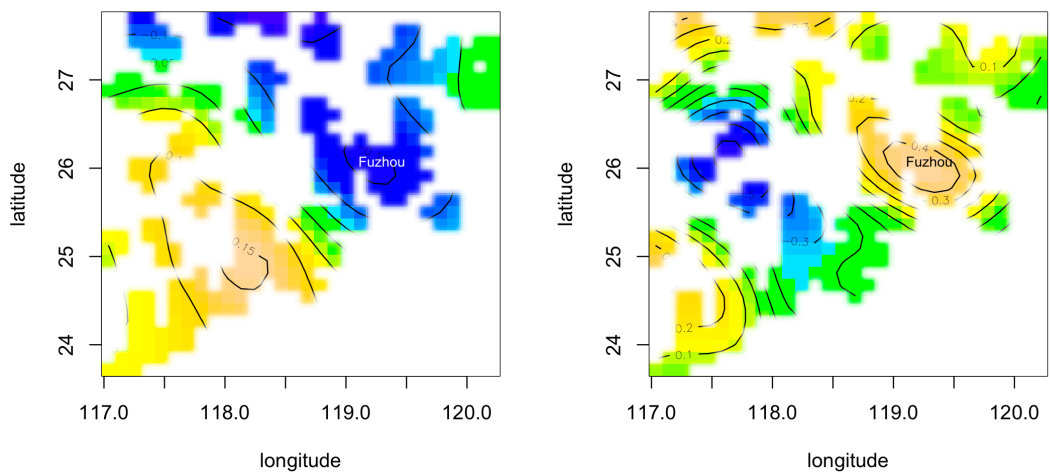

Figure 2: Contour plot of GAM results using a thin plate regression spline, where the response variable is phonotactic complexity (left) and average word length (right) of Min dialects, and the predictors, their latitude and longitude. Colder colors (blue) indicate low complexity, while warmer colors (yellow) indicate higher values.

Regarding studies on geographical distributions of linguistic complexity, Wichmann and Holman (2023) find that across 1,267 languages, average word length exhibits the behavior typical of typological linguistic features, showing sensitivity to areal influence detectable over a 5000 km range. Average word length is also significantly inversely correlated with (log) population size when compared across six macroareas, but inconsequential when measured within each macroarea.

Our study extends this line of typological research in linguistic complexity by focusing instead on a large number of closely-related language varieties, which has the benefit of holding many variables relatively constant in examining how word length interacts with other social or linguistic variables.

## 3. Dataset

Refer to Table 1 for a summary of the datasets we use in our work.

### 3.1. Dutch Dialect Data

Our data comes from the Goeman-Taeldeman-Van Reenen-Project (Taeldeman, Johan and Goeman, A, 1996), which is a Dutch dialect dataset collected between 1980 and 1995 in the Netherlands and Belgium. It contains phonetic transcriptions of 1,876 lexical items across 613 dialect sites elicited in a concept-aligned manner that enables fair comparison between sites.[2] This dataset has enjoyed widespread usage in dialectometry (Wieling et al., 2007), and the version we utilize in this study comes in a preprocessed form that is typical for dialectometric usage, where the diacritics are removed to reduce sparsity. Furthermore, a subset of 562 lexical items are selected to exclude morphological inflections of the same item, along with multiword expressions. Due to differences in transcription between the data in Belgium and in the Netherlands, we base our analysis on data from the latter country, which contains 424 sites (Wieling et al., 2007). After removing Frisian, a separate language that belongs to a different branch of West Germanic, we have 366 Dutch sites.

### 3.2. Min Dialect Data

Our Min data comes from Centre for the Protection of Language Resources of China (2023), which contains phonetic transcriptions for 1,000 characters and 1,200 concepts across 1,289 Sino-Tibetan varieties in mainland China and Taiwan.[3] Min is a subgroup of internally divergent Sinitic varieties. We only include the Min dialects spoken in

---
[2]The original dataset contains 1876 items, but we use Wieling et al. (2007)'s 562-item subset.
[3]Unfortunately, they do not allow the data to be publicly released. We obtained our version on June 12, 2023.

Fujian Province (where most Min dialects are concentrated).

We choose concept-aligned over character-aligned data for Min since the number of phones per character does not vary much due to the relatively constant syllable structure across Sinitic languages[4], and character-aligned data would artificially restrict the word length since concepts are not always expressed as single characters in Chinese. As with any concept-aligned data, this may mean that different Min dialect entries may not be cognate with each other due to lexical innovation or borrowing. Furthermore, for concept-aligned data, the elicitation process often obtains pronunciation variants that may or may not be related to one another within a site. For the purposes of this study, we pick the first variant, as our goal is to model the phonotactic sequence, and would arguably not lose essential information in keeping only one variant.

## 4. Methods

### 4.1. Phonotactic Language Modeling

We use Pimentel et al. (2020)'s unidirectional LSTM-based (Hochreiter and Schmidhuber, 1997) phoneme-level language model, which assigns probabilities to the phonetic transcriptions of words and learns phonotactically valid sequences of phones. A separate language model is learned for each dialect in our dataset. When the correlation between phonotactic complexity and word length is calculated, word length is measured in phonemes. To measure the phonotactic complexity of a language (or dialect in our case), Pimentel et al. (2020) estimate the Shannon entropy of the variety, which quantifies the average information encoded in each phoneme of a word (bits per phoneme). Entropy is an appropriate measure of phonotactic complexity, given the branching interpretation we provide in section 1. Because it is intractable to iterate over all possible strings to calculate the exact Shannon entropy, the authors use the average negative log-probability of a word to approximate the cross-entropy between the true, unknown phonotactic probability distribution of the language $p_{lex}$ and the one learned by the model $q_{lex}$. Cross-entropy in turn provides an upper bound on the true entropy:

$$H(p_{lex}) \leq H(p_{lex}, q_{lex}) \approx -\frac{1}{N} \sum_{i=1}^{N} \log q_{lex}(\tilde{\mathbf{x}}^{(i)}) \quad (1)$$

We refer the reader to their work for the full derivation.

---

[4]Each character is one syllable.

### 4.2. Phonotactics and Syllable Structure

Pimentel et al. (2020)'s phonotactic language model does not explicitly consider how the location of a phoneme or set of phonemes within a syllable can determine the behavior of the phoneme. Onsets, for example, are often the locus of phonemic contrast, while codas are often absent or are the location of positional neutralization (Hayes, 2008). Phonological rules often apply at such syllable edges (Blevins, 1995). Syllable constituents—the onset, nucleus, and coda—also differ in the types of complex (multi-segment) sequences allowed (Blevins, 1995). Thus an accurate account of phonotactic constraints and thus phonotactic complexity must account for syllable structure. If we measure phonotactic complexity by the number of options at some position in a phonethen how many options are next depends on what syllable constituent the next phone occupies.

### 4.3. Syllabification

To obtain syllable constituents for Dutch, we use `syllabiphon`[5]. This Python package syllabifies phoneme sequences according to the Sonority Sequencing Principle (Selkirk, 1984), which states that—across different languages—syllables rise in sonority until a peak, after which the sonority falls. We first identify syllable boundaries and then split each syllable into onset (which is level or rising in sonority), nucleus (the sonority peak), and coda (the remainder of the syllable, which remains level or falls in sonority). For example, [ʔɒrdə] (*aarde* in the Aalsmeer NH dialect) is syllabified as:

[Syl(ons='ʔ', nuc='ɒ', cod='r'),
 Syl(ons='d', nuc='e', cod='')]

Sonority scores for each phone are obtained from PanPhon and are on a scale from 1–9 (9 being the most sonorous) (Mortensen et al., 2016). Vowels have minimum sonority 8.

We postprocess the result of syllabiphon to merge diphthongs and triphthongs into the nuclei of their syllables. On a set of one randomly chosen word from 100 random Dutch dialects, syllabiphon, with the aforementioned diphthong/triphthong postprocessing, correctly syllabified 85% of the items upon manual inspection. Almost all of the errors concern consonant clusters involving [s] (e.g. [sp], [st], [sx]) being split into different constituents.[6] ML-based syllabifica-tion models achieve higher performance on individual languages. Syllabiphon has the advantage, though, of performing reasonably well on IPA transcriptions of any language without supervision.

---

[5]https://anonymous.4open.science/r/syllabiphon-413E/

[6]Since [s] has a lower sonority when it is in a word-initial or word-final cluster with a stop (e.g. [sp], [ps]), we manually set the sonority of [s] in these positions to be 0 for Dutch to ensure the cluster is placed into the

| Language | Source | # dialects (we use) | # words |
|---|---|---|---|
| Dutch | Taeldeman, Johan and Goeman, A (1996) | 366 | 562 |
| Min | Centre for the Protection of Language Resources of China (2023) | 60 | 1200 |

Table 1: Statistics on the datasets in our experiments

As for Min, the data is already implicitly segmented by syllable since each Han character (which comes with its own tone) is monosyllabic in Sinitic, so we use the tone numbers as syllable boundaries. To identify syllable constituents, we assign all phones with a [-cons] feature from PanPhon (Mortensen et al., 2016) (except for ʔ) to the nucleus. Phones with [+cons] (and ʔ) are assigned to the onset when no nucleus has been identified for the syllable yet. Additionally, contour tones are treated as one token. To summarize, 'ŋi31tʰœ51' is syllabified as [Syl('ŋ', 'i', '', '31'), Syl('tʰ', 'œ', '', '51')].

### 4.4. The Syllable Constituency Prediction Task

To incorporate syllable constituency into our model, we add the objective of predicting the syllable constituents **c** of a word given its phonetic transcription **x**:

$$p(\mathbf{c}|\mathbf{x}) = \prod_{i=1}^{|x|} p(c_i|\mathbf{x}_{\leq i}) \quad (2)$$

In doing so, the model learns about the distribution of phonemes in different syllable constituents—a key aspect of phonotactics as described earlier (subsection 4.2). We first feed the phoneme sequence in, apply an embedding layer to each phone, and feed the embedding sequence into the LSTM. The output of the LSTM is used for both the language modeling task and the auxiliary task of syllable constituency prediction. Specifically, at each step, we predict both the next phone and the next phone's syllable constituent (Figure 3).

Note that we only include the phone-level language modeling loss in the calculation of the correlation, so we model the same distribution as Pimentel et al. (2020), which is the classic language modeling objective (reproduced from their paper):

$$p(\mathbf{x}) = \prod_{i=1}^{|x|} p(x_i|\mathbf{x}_{<i}) \quad (3)$$

---
same constituent. This changed the accuracy to 93%.

We cannot feed the syllable constituent as input to the language modeling prediction task because the syllable constituent is determined using global information, which autoregressive language modeling should not have access to. Doing so would artificially increase the predictability of the phoneme (and thus deflate the phonotactic complexity). A longer sequence may be more predictable (and thus have lower complexity) because it may have more syllables and thus more global information.

### 4.5. Multi-task Loss

To jointly optimize for both phone and constituency prediction, a simple strategy would be to simply sum up the losses, which are each weighted by hyperparameters $\lambda_{a,b}$:

$$\mathcal{L}(\mathbf{x}, \mathbf{c}) = \lambda_a \mathcal{L}_{phon}(\mathbf{x}) + \lambda_b \mathcal{L}_{syl}(\mathbf{x}, \mathbf{c}) \quad (4)$$

where $\mathcal{L}_{phon}(\mathbf{x})$ is the cross entropy version of the likelihood $p(\mathbf{x})$, where $\mathcal{L}_{syl}(\mathbf{x}, \mathbf{c})$ is the cross entropy version of the likelihood $p(\mathbf{c}|\mathbf{x})$, and where, again, only $\mathcal{L}_{phon}(x)$ is used for the correlation.

However, Kendall et al. (2018) show that the performance of multi-task models is highly dependent on an appropriate choice of weighting between each task's loss, which may be difficult to decide on with manual selection. We therefore follow Kendall et al. (2018) in opting for dynamically weighing the losses by assigning less weight to more uncertain tasks, where uncertainty is computed here with the variance of the task-specific losses, $\sigma_t^2$:

$$\tilde{L}_t = \frac{1}{2\sigma_t^2} L_t + \log \sigma_t \quad (5)$$

where $\log \sigma_t$ here serves to avoid division by zero. Such adjustments have improved multitask learning in various NLP tasks (Hofmann et al., 2023; Hung et al., 2023). In practice, given the instability in regressing the variance $\sigma_t^2$, we train the network to predict the log variance $\eta_t := \log \sigma_t^2$:

$$\tilde{L}_t = \frac{1}{2}(e^{-\eta_t} L_t + \eta_t) \quad (6)$$

The final loss is then the sum of the uncertainty-adjusted losses over all tasks.

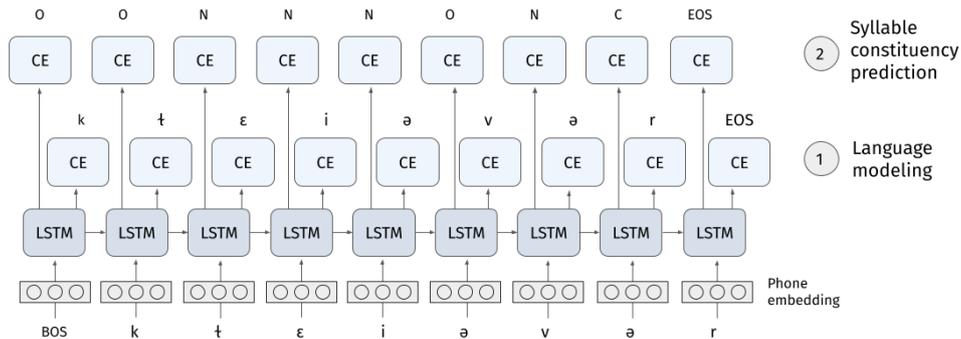

Figure 3: Diagram of our phonotactic language model with syllable constituency prediction as an auxiliary task, where O means onset; N, nucleus; C, coda; CE, cross-entropy loss. The example shown is the pronunciation of "klaver" in the Deurne NB dialect. The triphthong [ɛiə] was merged into the nucleus during postprocessing, but we treat diphthongs and triphthongs as sequences of multiple phonemes when calculating word length. The same LSTM is used for both language modeling and for syllable constituency prediction.

### 4.6. Generalized Additive Models

Having obtained approximations of phonotactic complexity, we use generalized additive models (GAM) (Wood, 2011, 2017) to model the scores as a function of their geographical coordinates. GAMs can be considered as an extension of generalized linear models (GLM) (McCullagh and Nelder, 1989). In a GLM, the mean $\mu$ of a random variable $Y$ is related to a weighted sum of linear predictors $X$ with coefficients $\beta$ (where $X\beta$ is denoted $\eta$) through a link function $g$:

$$g(\mu(Y)) = \eta = X\beta \quad (7)$$

In contrast to linear regression, the link function in the GLM provides the possibility of relating a linear model to an $\eta$ of a distribution in the exponential family. GAMs extend this by fitting functions on the predictors:

$$\eta = b_0 + f(x_1) + f(x_2) \ldots + f(x_p) \quad (8)$$

The functions are smoothing splines that are learned to fit to the predictors, and can take any number of arguments. The smoothing splines are themselves linear combinations of basis functions, which are geometrically simpler functions that combine by way of a weighted sum to model the data. The non-parametric nature of the smoothing splines allows for much greater flexibility in fitting the distribution of the variable, while the additivity of the model allows for understanding the contribution of each predictor, as the fitted variables do not depend on one another. In this paper, we use the implementation of GAMs offered in the *mgcv* package,[7] with thin plate regression splines as smooths (Wood, 2003). Our statistical results are shown in Table 3 and Table 4.

## 5. Results and Discussion

### 5.1. Phonotactic Complexity and Word Length

As shown in Table 2, we observe a moderate negative correlation across both Dutch and Min dialects. For the Dutch dataset, we observe a Pearson correlation of $-0.68$ both with and without syllabification, whereas for Min we observe a decrease from $-0.72$ to $-0.7$ after syllabification. The results show that incorporating the auxiliary task of syllable constituency does not strengthen the negative correlation for our datasets. This suggests that the LSTM-based phonotactic language model already implicitly learns syllable structure based on a phoneme's context and position within a word. Despite modifying the setup with a syllable structure-based approach, we still reach a similar negative correlation, attesting to the robustness of the correlation. Furthermore, we also note that the tradeoff between linguistic complexities may in fact extend beyond such a binary model, where multiple factors may be at play in addition to phonotactic complexity and word length (Levshina, 2022). As such, the correlation should not be seen as an evaluation metric, since it is not known whether there is indeed a complete tradeoff between just the two dimensions.

### 5.2. GAMs

Table 3 and 4 summarize the statistical results of our GAMs on the Dutch dialect dataset, where for both phonotactic complexity and average word length, the coordinates as predictors are found to be statistically significant [8]. The regression surfaces are visualized in Figure 1 as contour plots mapped upon the coordinates of the Netherlands, where the color blue indicate that the values are low, yellow indicating that they are high, and green being a transitional value between the two. We also include statistics of the raw values for the Dutch dataset in Figure 4 for comparison. Values in both GAMs can be seen to show a high degree of similarity with nearby points, which corresponds with the fact that dialects which are spoken near one another tend to be more similar to one another than those that are spoken far away (Nerbonne, 2010). The larger blue region in the phonotactic complexity plot of Figure 1 is the region of Holland. For Min, high average word length and low phonotactic complexity is similarly concentrated around the capital of Fuzhou (Figure 2).

---
[7] https://cran.r-project.org/web/packages/mgcv/index.html

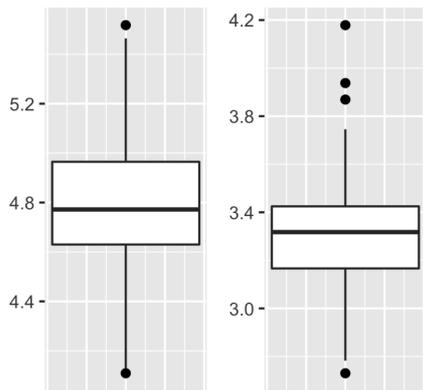

Figure 4: Box plot of average word length (left) and phonotactic complexity (right) in Dutch dialects.

In the remaining sections, we focus our discussion on the Dutch dialect dataset due to a larger quantity of Dutch dialectology studies that examine this topic, although similar trends are observed also in the Min dataset.

### 5.3. The Compensation Hypothesis

Our results in Table 2 affirm that the compensatory relationship between phonotactic complexity and average word length observed in Pimentel et al. (2020) applies to the level of dialects as well. Dialects, however, provide us with a spatial dimension with social and historical correlates against which we can compare linguistic variables. We make use of this fact by modelling the phonotactic complexity and average word length of dialects as functions of their coordinates. In ob-serving the inverse relationship between the distribution of high and low values of Figure 1, we show how the negative correlation appears on a geographical level.

---

[8] All GAM results including the plots are for non-syllabified runs.

### 5.4. Koineization and the Linguistic Niche Hypothesis

In observing the center of low phonotactic complexity and high average word length to concentrate in the region of Holland, we interpret this as potential evidence of the linguistic niche hypothesis (Lupyan and Dale, 2010; Dale and Lupyan, 2012), which states that linguistic complexity adapts to social constraints (Trudgill, 2001; McWhorter, 2007; Bentz and Winter, 2014). Lupyan and Dale (2010) conduct a study of over 2000 languages, and found morphological complexity to be negatively correlated with population size, the degree of language contact, and the size of the region—the three of which are correlated—as languages spoken throughout regions with larger populations and broader geographical extents would also have a higher likelihood of having many non-native speakers. Extensive contact of this kind has been suggested to induce grammatical simplification, as non-native adult learners of the language would have difficulty in acquiring morphologically complex languages compared to a child.

As for dialects, Dutch dialectology has long utilized similar arguments, where the large influx of immigrants to early modern Holland is claimed to be the central cause of significant grammatical simplification in the urban vernacular of Holland (Hendriks et al., 2018; Howell, 2006), the effect of which remains observable in its modern form (Kloeke, 1927; Hamans, 2011). Holland in the 16th to 18th century saw a massive number of immigrants. Of 1.2 million immigrants, only over 200,000 were from within the province of Holland. The remainder came mostly from southern Netherlands and Germany (Lucassen, 2002; Hendriks et al., 2018). This created a situation in which speakers of many different Dutch varieties intermixed (since natives of the province had become a minority). Moreover, in an effort to integrate and achieve better socioeconomic outcomes, linguistic forms which carry strong regional characteristics were avoided; there was a bias towards phonologically and lexically simple forms due to the learning constraints of adult speakers. This led all parties to converge on simplified linguistic compromise. Children in such multi-dialectal contexts also have to make sense of the language used around them and, in the process, simplify the wide range of dialectal input in their environment. The combination of the aforementioned fac-tors is claimed to have established a simpler variety, which became the native language of further generations (Howell, 2006; Kerswill and Williams, 2000).

| Language | Syllable structure | Pearson *r* | Spearman *ρ* |
|---|---|---|---|
| Dutch | No | -0.678897 | **-0.63107** |
| Dutch | Yes | **-0.684041** | -0.627137 |
| Min | No | **-0.720228** | -0.692592 |
| Min | Yes | -0.698437 | -0.665027 |

Table 2: Correlation between phonotactic complexity and word length with and without syllable constituency prediction.

| Component | Term | Estimate | Std Error | t-value | p-value |
|---|---|---|---|---|---|
| parametric coefficients | Intercept | 3.299 | 0.007 | 498.1 | <0.001 |
| **Component** | **Term** | **edf** | **Ref. df** | **F-value** | **p-value** |
| smooth terms | s(longitude, latitude) | 25.4 | 28.22 | 18.12 | <0.001 |
| **Adjusted R-squared** | | | | | |
| 0.581 | | | | | |

Table 3: Statistical results of GAMs on phonotactic complexity in Dutch dialects.

| Component | Term | Estimate | Std Error | t-value | p-value |
|---|---|---|---|---|---|
| parametric coefficients | Intercept | 4.805 | 0.009 | 524.4 | <0.001 |
| **Component** | **Term** | **edf** | **Ref. df** | **F-value** | **p-value** |
| smooth terms | s(longitude, latitude) | 24.85 | 27.98 | 15.57 | <0.001 |
| **Adjusted R-squared** | | | | | |
| 0.543 | | | | | |

Table 4: Statistical results of GAMs on average word length in Dutch dialects.

In our results, Figure 1 is compatible with both the Linguistic Niche hypothesis and the Compensation Hypothesis—phonotactics were simplified through koineization and word length increased concomitantly. As for Min, despite a relative lack of literature on linguistic complexity, the concentration of high average word length and low phonotactic complexity near the capital of Fuzhou may arguably also be due to similar processes (Figure 2), given the greater amount of socioeconomic opportunities offered.

Although Shcherbakova et al. (2023) find no inverse correlation between morphosyntactic complexity and sociodemographic factors such as the proportion of L2 speakers when examined against a database of 1314 languages (Skirgård et al., 2023), our study differs from theirs both in the dimensions of linguistic complexity measured, and also in the scale of linguistic variation, where our emphasis is placed on microvariation over geographically proximate dialects. As such, our results do not necessarily stand in conflict with theirs given the differences in measurement, and may point to the need of a sharper formulation of the Linguistic Niche hypothesis for more precise predictions of how linguistic complexity and sociodemographic factors affect one another.

## 6. Future Work

Although we have based this study on two languages of different language families to ensure the robustness of the results across typologically different languages, we seek to validate our results on additional dialect datasets, for which an interesting candidate would be the *Phonetischer Atlas von Deutschland* Göschel (2000),[9] a German dataset containing 29,530 words transcribed in IPA across 183 cities in Germany.

We also seek to explore different methods of incorporating syllable structure beyond learning syllable constituency as an auxiliary task. As Daland et al. (2011a) suggest, we can also explicitly supervise the model with sonority. Furthermore, despite Pimentel et al. (2020)'s finding that phonological features do not improve performance, we can explore more sophisticated methods of incorporating such features, as Torre (2003) finds that place of articulation is important in Dutch phonotactics. Finally, despite the lack of increase in negative correlation, we seek to perform a more thorough evaluation of our model enriched with syllable constituency knowledge on other downstream tasks.

---
[9] https://github.com/cysouw/PAD

## Limitations

Our study is limited by the fact that the dataset we use comes in a form where diacritics are removed, which could potentially encode phonemic distinctions, thereby decreasing entropy. Additionally, our unidirectional LSTM limits our capability to capture regressive assimilation, which we hope to improve upon with more expressive architectures such as Transformers (Vaswani et al., 2017) in future work, though we acknowledge Mayer and Nelson (2020)'s suggestion that Elman RNNs may suffice to model most phonotactics. We also acknowledge that some phones may be inherently more complex than others due to differences in articulatory difficulty (Maddieson, 2009), which our model does not capture. Finally, we reduce the factors behind linguistic complexity to a binary tradeoff between two dimensions, but Levshina (2022) suggest the situation is actually multivariate.

## 7. Bibliographical References

## 8. Language Resource References